\documentclass{article}

\usepackage{pstricks,pst-grad}
\usepackage{graphicx}
\usepackage{amsmath}
\usepackage{psfrag}
\usepackage{amssymb}
\usepackage{url}

\numberwithin{equation}{section}

\newcommand {\R}{\mathbb {R}}

\newcommand {\argmax}{{\rm arg}\max\limits}

\newcommand {\argmin}{arg\min\limits}

\begin{document}
  \begin{center}
    {\Large  Convolutional Matching Pursuit and Dictionary Training}
    \bigskip
    
    {\large Arthur Szlam, Koray Kavukcuoglu, and Yann LeCun}
  \end{center}

\section{Introduction}
  One of the most succesful recent signal processing paradigms has been the sparse coding/dictionary design model \cite{olshausen97sparse,bruckstein:34}. In this model,       
we try to represent a given $d\times n$ data matrix $X$ of $n$ points in $\R^d$ written as columns via a solution to the  problem                                  
\[\{W_*,Z_*\}=\{W_*(K,X,q),Z_*(K,X,q)\}\]                                                                                                                                 
\begin{equation}=\argmin_{Z\in\R^{K\times n},W\in \R^{d\times K}} \sum_k||Wz_k-x_k||^2, \, \, ||z_k||_0\leq q,\label{l0}\end{equation}                                    
or its $Z$ coordinate convexification                                                                                                                                     
\[\{\tilde{W}_*,\tilde{Z}_*\}=\{\tilde{W}_*(K,X,\lambda),\tilde{Z}_*(K,X,\lambda)\}\]                                                                                     
\begin{equation}=\argmin_{Z\in\R^{K\times n},W\in \R^{d\times K}} \sum_k||Wz_k-x_k||^2+\lambda||z_k||_1.\label{l1}\end{equation}                                          
Here, 
$\{W,Z\}$ are the dictionary and the  coefficients, respectively, and $z_k$ is the $k$th column of $Z$.   $K$, $q$, and $\lambda$ are user selected parameters controlling the power of the model.

More recently, many models with additional structure have been proposed.  For example, in \cite{Yuan06modelselection,jmlrBach08a}, the dictionary elements are arranged in groups and the sparsity is on the group level.  
In \cite{Behnkeconvnnmf,koraynips2010,zeilercvpr2010}, the dictionaries are constructed to be translation invariant.  In the former work, the dictionary is constructed via a non-negative matrix factorization.  In the latter two works, the construction is a convolutional analogue of \ref{l1} or an $l^p$ variant, with $0<p<1$.  
In this short note we work with greedy algorithms for solving the convolutional analogues of \ref{l0}.  Specifically, we demonstrate that sparse coding by matching pursuit and dictionary learning via K-SVD \cite{elad-ksvd} can be used in the translation invariant setting.

\section{Matching Pursuit}
Matching pursuit \cite{Mallat93matchingpursuit} is a greedy algorithm for the solution of the
sparse coding problem
\[\min_z ||Wz-x||^2, \]\[ ||z||_0\leq q,\]
where the $d\times k$ matrix $W$ is the dictionary, the $k\times 1$
$z$ is the code, and $x$ is an $d\times 1$ data vector.

\begin{enumerate}
\item Set $e=x$, and $z$ the $k$-dimensional zero vector.
\item Find $j=\argmax_i ||W_i^Te||_2^2$.
\item Set $a=W_j^Tx$.
\item Set $e\leftarrow e-aW_j$, and $z_j=z_j+a$.
\item Repeat for $q$ steps
\end{enumerate}

Note that with a bit of bookkeeping, it is only necessary to
multiply $W$ against $x$ once, instead of $q$ times.  This at a cost
of an extra $O(K^2)$ storage:
set $e_r$ and $a_r$ be $e$ and $a$ from the $r$th step above.  Then:
\[W^Te_0=W^Tx;\]
\[W^Te_1=W^Tx-a_0W^TW_{j_0},\]
and so on.  If the Gram matrix for $W$ is stored, this is just a lookup.
\subsection{Convolutional MP}

We consider the special case
\[\min_z ||\sum_{j=1}^k w_j*z_j-x||^2,\]\[||\overline{z}||_0\leq q,\]
where each $w_j$ is a filter, and $\overline{z}$ is all of the
responses.

Note that the Gram matrix of the ``Toeplitz'' dictionary consisting
of all the shifts of the $w_j$ is usually too big to be used as a
lookup table. However, because of the symmetries of the convolution,
it is also unnecessary; we only need store a $4*h_f\times w_f\times
k^2$ array of inner products, where $h_f$ and $w_f$ are the
dimensions of the filters.

With this additional storage, to run $q$ basis pursuit steps with
$k$ filters on an $h\times w$ image costs the computation of one
application of the filter bank plus $O(kqhw)$ operations.

\section{Learning the filters}

Given a set of $x$, we can learn the filters and the codes simultaneously.  
Several methods are available.  A simple one is to alternate
between updating the codes and updating the filters, as in K-SVD
\cite{elad-ksvd}:

\begin{enumerate}
\item Initialize $k$ $h_f\times w_f$ filters $\{w_1,...,w_k\}$.
\item Solve for $z$ as above.
\item For each filter $w_j$,
\begin{itemize}
\item find all locations in all the data images where $w_j$
is activated
\item extract the $h_f\times w_f$ patch $E_p$ from the reconstruction via
$z$ at each activated point $p$.
\item remove the contribution of $w_j$ from each $E_p$ (i.e. $E_p
\leftarrow E_p-c_{(p,j)}w_j$, where $c_{(p,j)}$ was the activation
determined by $z$).
\item update $w_j\leftarrow \text{PCA}(E_p)$
\end{itemize}
\item Repeat from step 2 until fixed number of iterations.
\end{enumerate}

We note that the forward subproblem (finding $Z$ with $W$ fixed) is
not convex, and so the alternation is not guaranteed to decrease the
energy or to converge to even a local minimum.  However, in
practice, on image and audio data, this method generates good filters.

\section{Some experiments}

We train filters on three data sets: the AT\&T face database, the motorcycles from a Caltech database, 
and the VOC PASCAL database.
For all the images in all our experiments, we perform an additive contrast normalization:
each image $x$ is transformed into $x'=x-x*b$, where $b$ is a $5\times 5$
averaging box filter.  This is very nearly transforming $x'=\nabla^2x$, that is,
using the discrete Laplacian of the image instead of the image.
Using the Laplacian would correspond to using the energy
\[\sum_x ||\nabla\left(\sum_j w_j*z_j-x\right)||^2,\]
that is, the energy sees the difference between gradients, not
intensities.

\subsection{Faces}
\begin{figure}
\begin{center}
\begin{minipage}{0.4\textwidth}
\includegraphics[width=.8\textwidth]{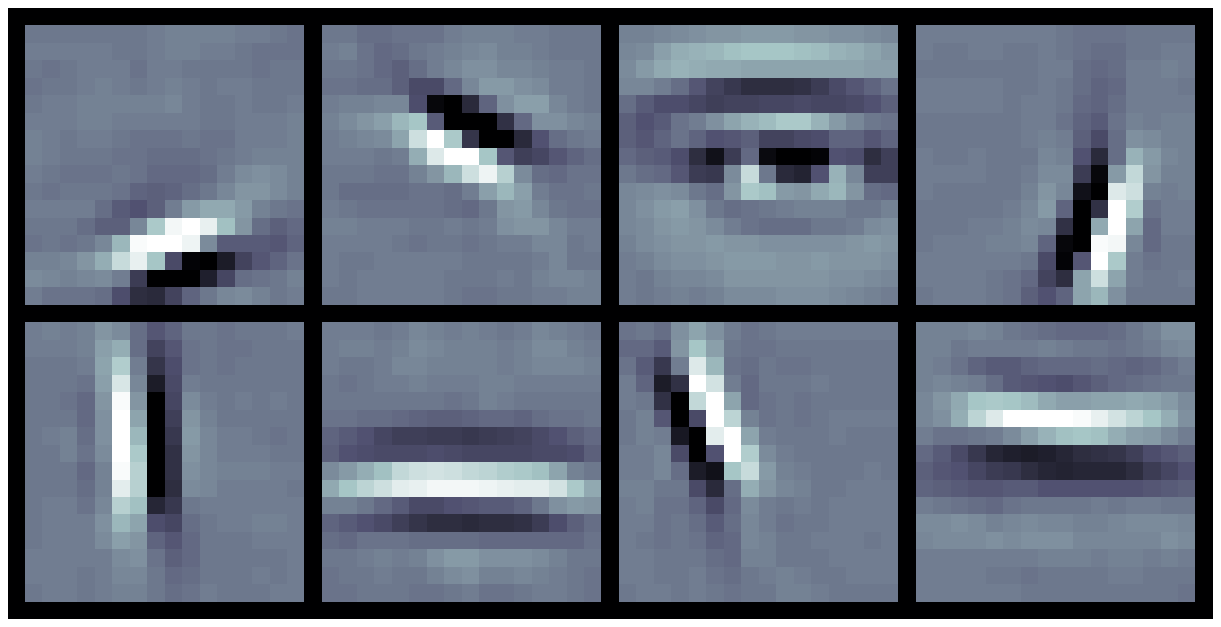}
\end{minipage}
\begin{minipage}{0.4\textwidth}
\includegraphics[width=.8\textwidth]{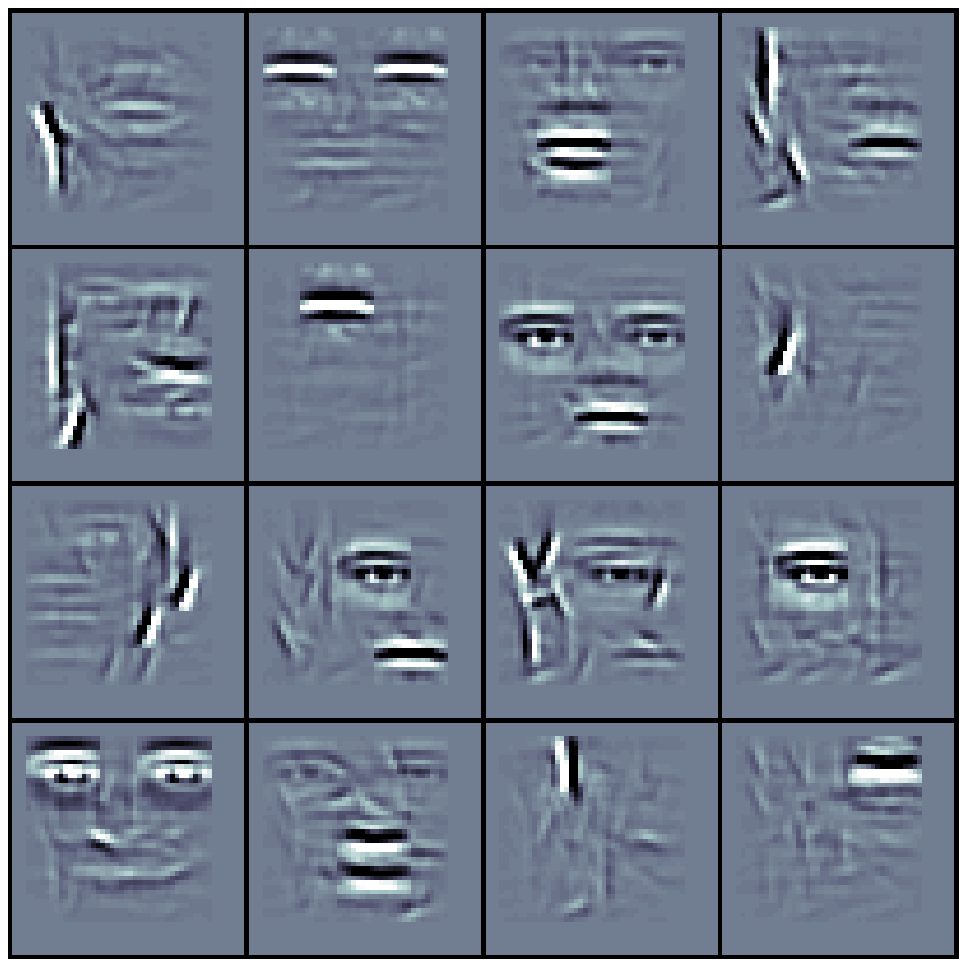}
\end{minipage}
\end{center}
\caption{First and second layer filters from faces}
\label{facefig}
\end{figure}

The AT\&T  face database, available at \url{http://www.cl.cam.ac.uk/research/dtg/attarchive/facedatabase.html} is a set of 400 images of 40 individuals.  The faces are centered in each image.  We
resize each image to $64\times 64$ and contrast normalize.  We train $8$ $16\times 16$ filters.  After training the filters we find the feature maps of 
each image in the database, obtaining a new set of 400 $8$ channel images.  We take the elementwise absolute value of each 
of the 8 channel images, and then average pool over $8\times 8$ blocks.   We then train a new 16 element dictionary on the 
subsampled images.  In figure \ref{facefig} we display the first level filters, and the second level filters up to shifts of size 8 and sign changes of the first level filters..

\begin{figure}
\begin{center}
\begin{minipage}{0.4\textwidth}
\includegraphics[width=.8\textwidth, height=.8\textwidth]{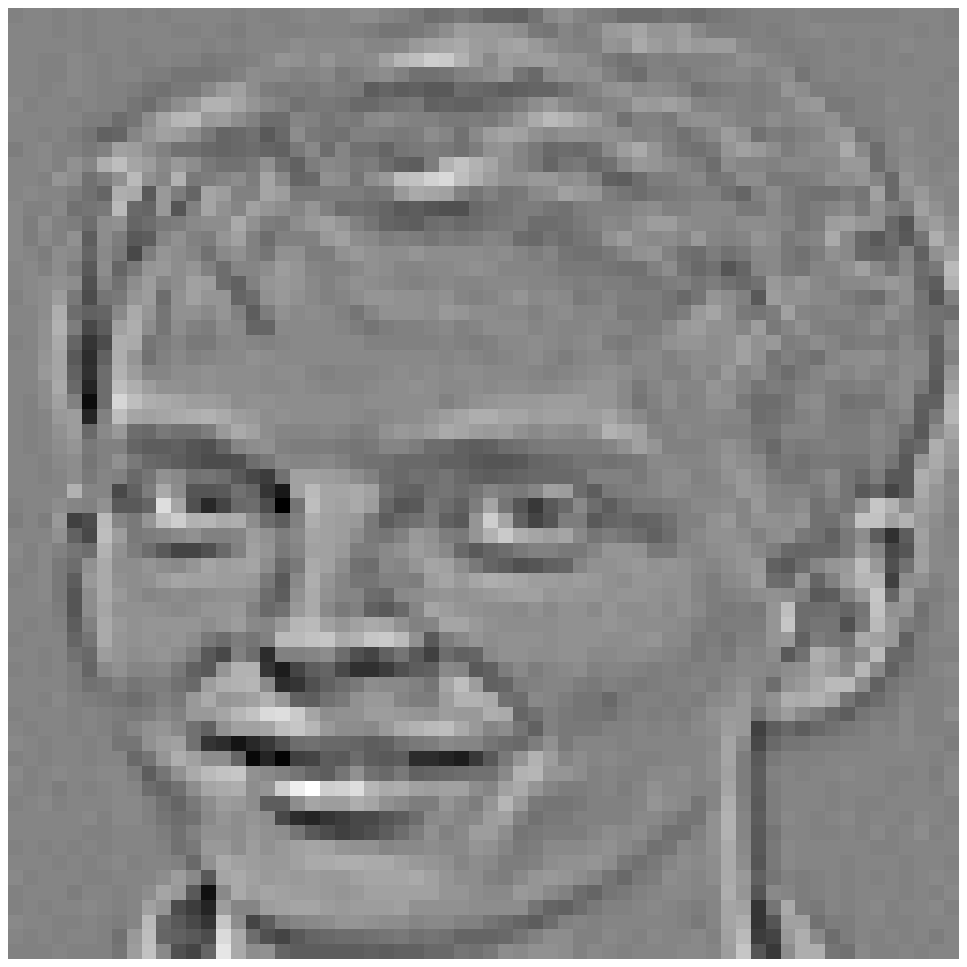}
\end{minipage}
\begin{minipage}{0.4\textwidth}
\includegraphics[width=.8\textwidth, height=.8\textwidth]{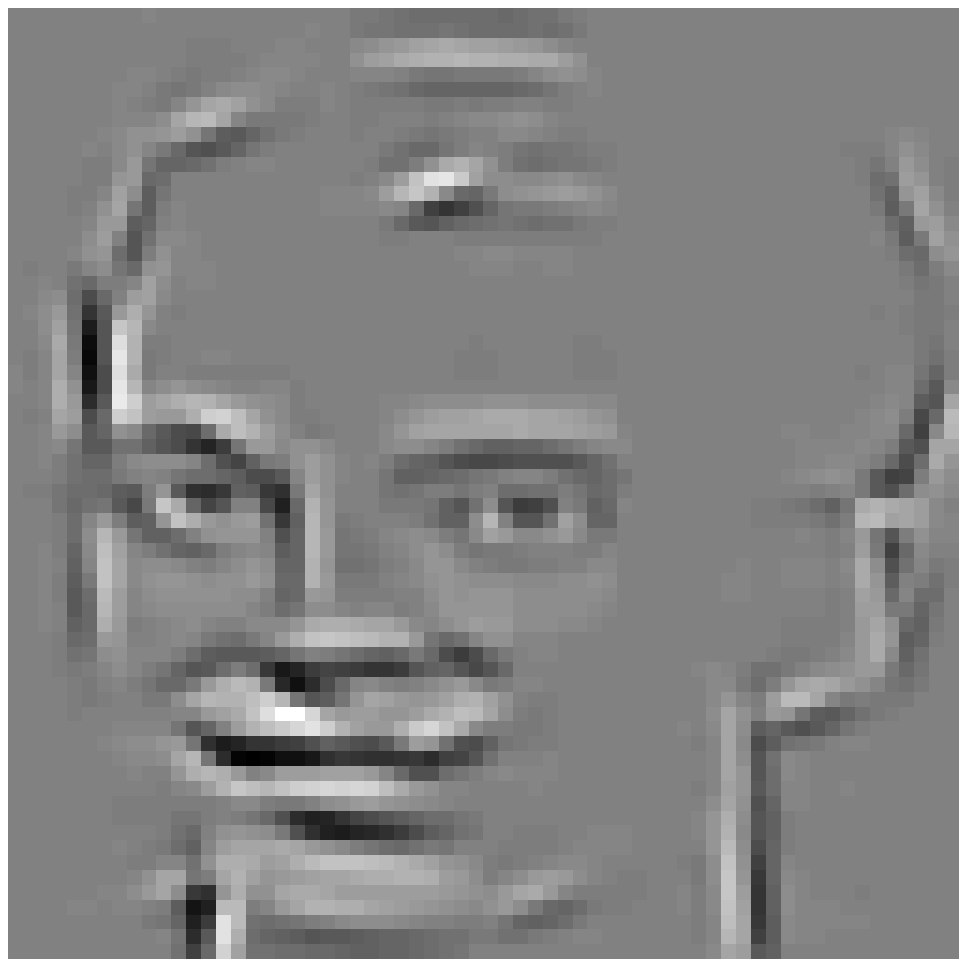}
\end{minipage}
\end{center}
\caption{a contrast normalized face, and its reconstruction
from 40 filter responses.}
\end{figure}

\subsection{Caltech motorcycles}
\begin{figure}
\begin{center}
\begin{minipage}{0.4\textwidth}
\includegraphics[width=.8\textwidth]{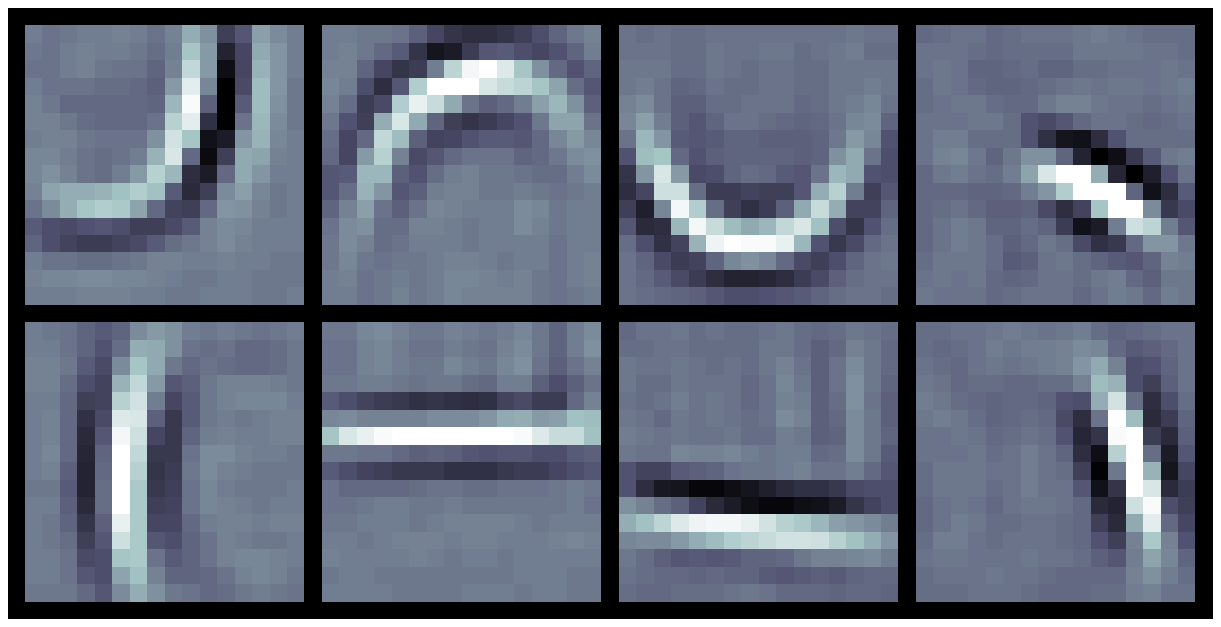}
\end{minipage}
\begin{minipage}{0.4\textwidth}
\includegraphics[width=.8\textwidth]{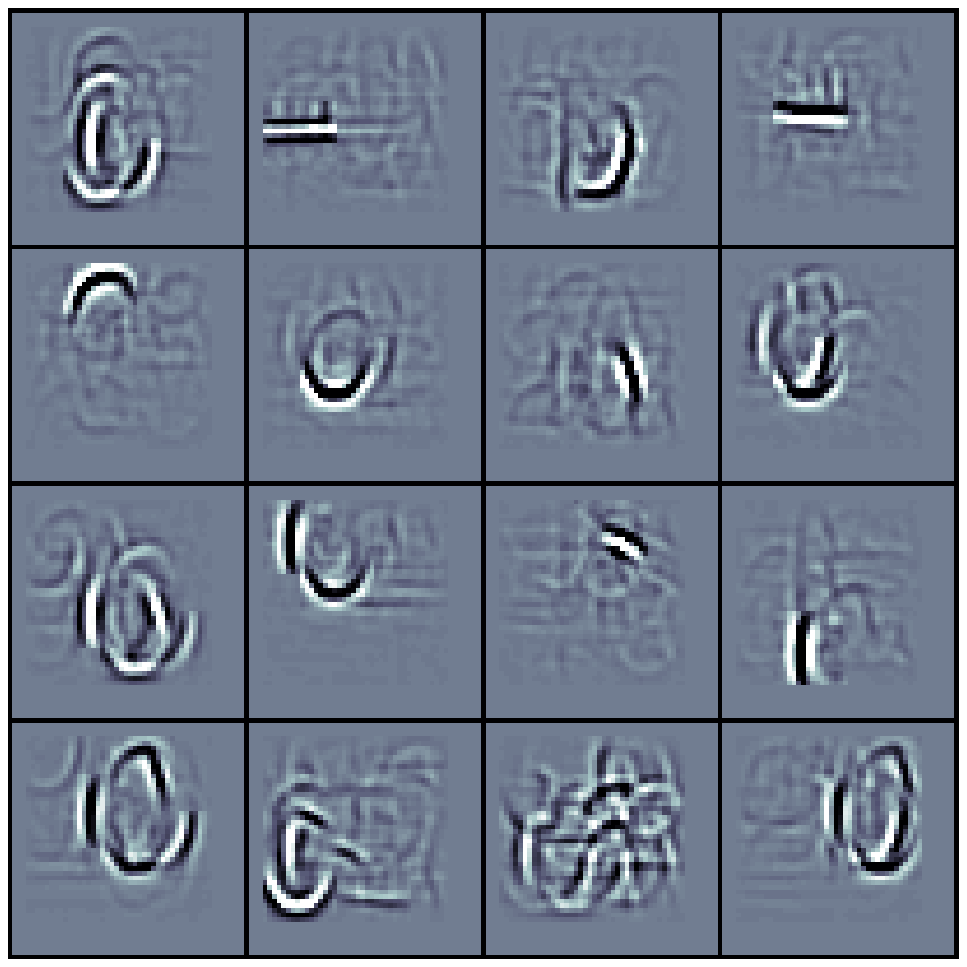}
\end{minipage}
\end{center}
\caption{First and second layer filters from motorcycles}
\label{bikefig}
\end{figure}

We also train on the motorbikes-side dataset, available at \url{http://www.vision.caltech.edu/html-files/archive.html} which consists of color images of various motorcycles.  The motorcycles are centered in each image.  
We convert each image to gray level, resize to $64\times 64$, and contrast normalize.  We train $8$ $16\times 16$ filters.     As before, we then train a new 16 element dictionary on the 
subsampled absolute value rectified responses of the first level.  In figure \ref{bikefig} we display the first level filters, and the second level filters up to shifts of size 8 and sign changes of the first level filters..

\begin{figure}
\begin{center}
\begin{minipage}{0.4\textwidth}
\includegraphics[width=.8\textwidth, height=.8\textwidth]{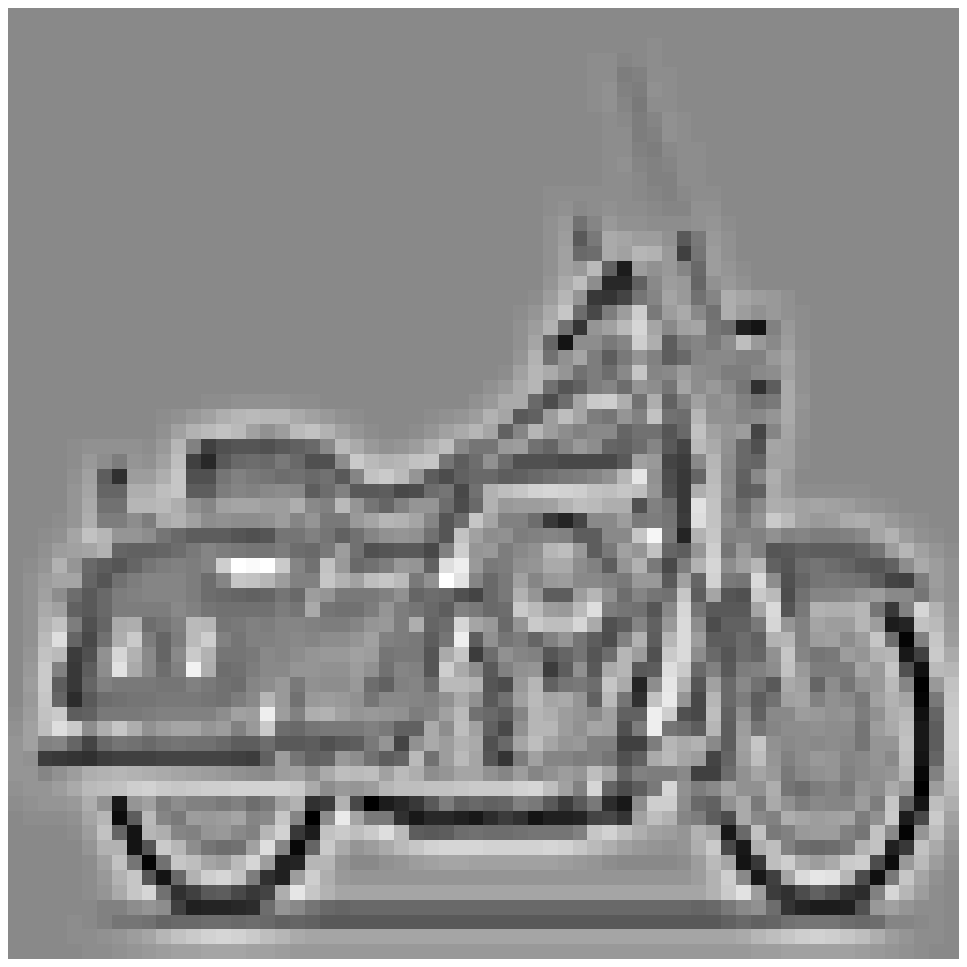}
\end{minipage}
\begin{minipage}{0.4\textwidth}
\includegraphics[width=.8\textwidth, height=.8\textwidth]{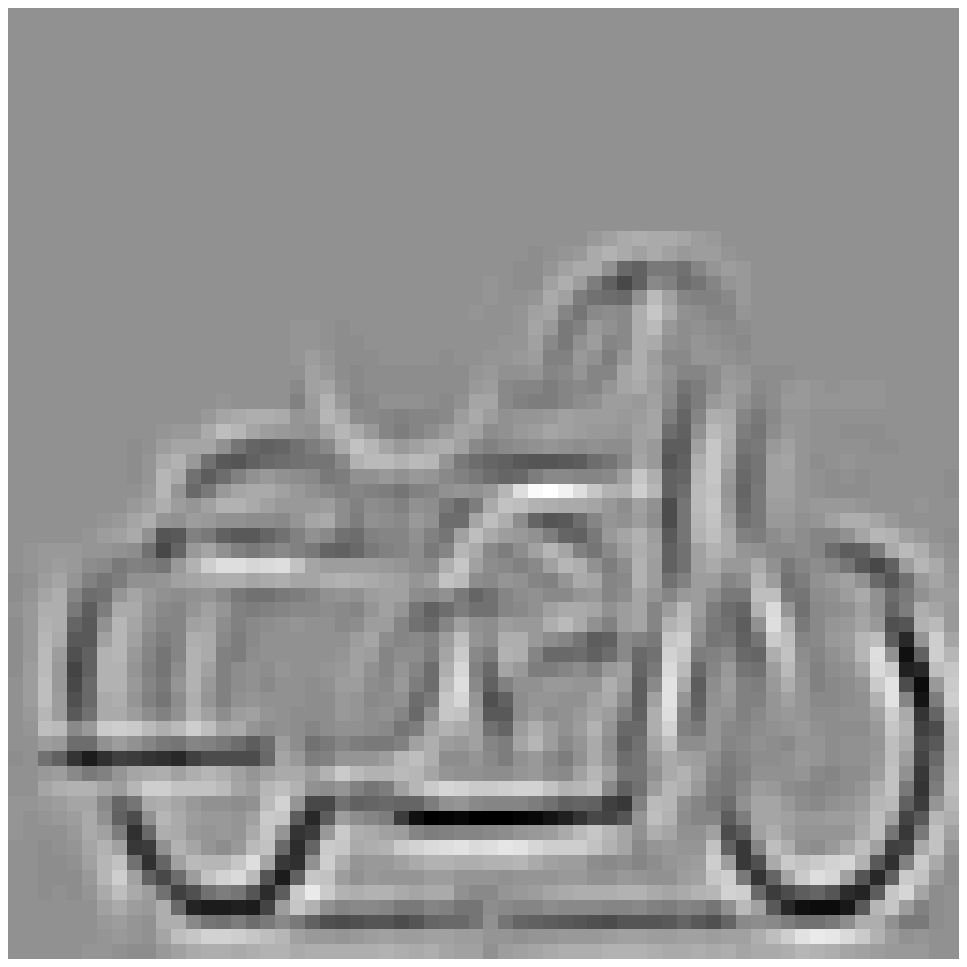}
\end{minipage}
\end{center}
\caption{A contrast normalized motorcycle, and its reconstruction
from 40 filter responses.}
\end{figure}

\subsection{Images from PASCAL VOC}
\begin{figure}
\begin{center}
\begin{minipage}{0.4\textwidth}
\includegraphics[width=.8\textwidth]{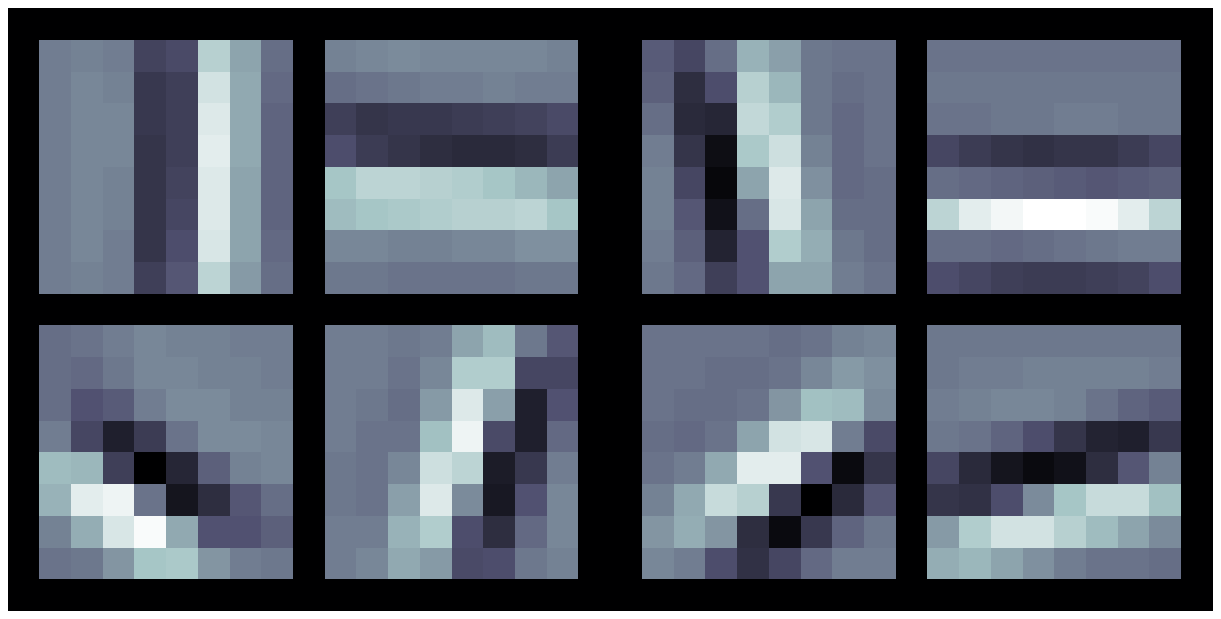}
\end{minipage}
\begin{minipage}{0.4\textwidth}
\includegraphics[width=.8\textwidth]{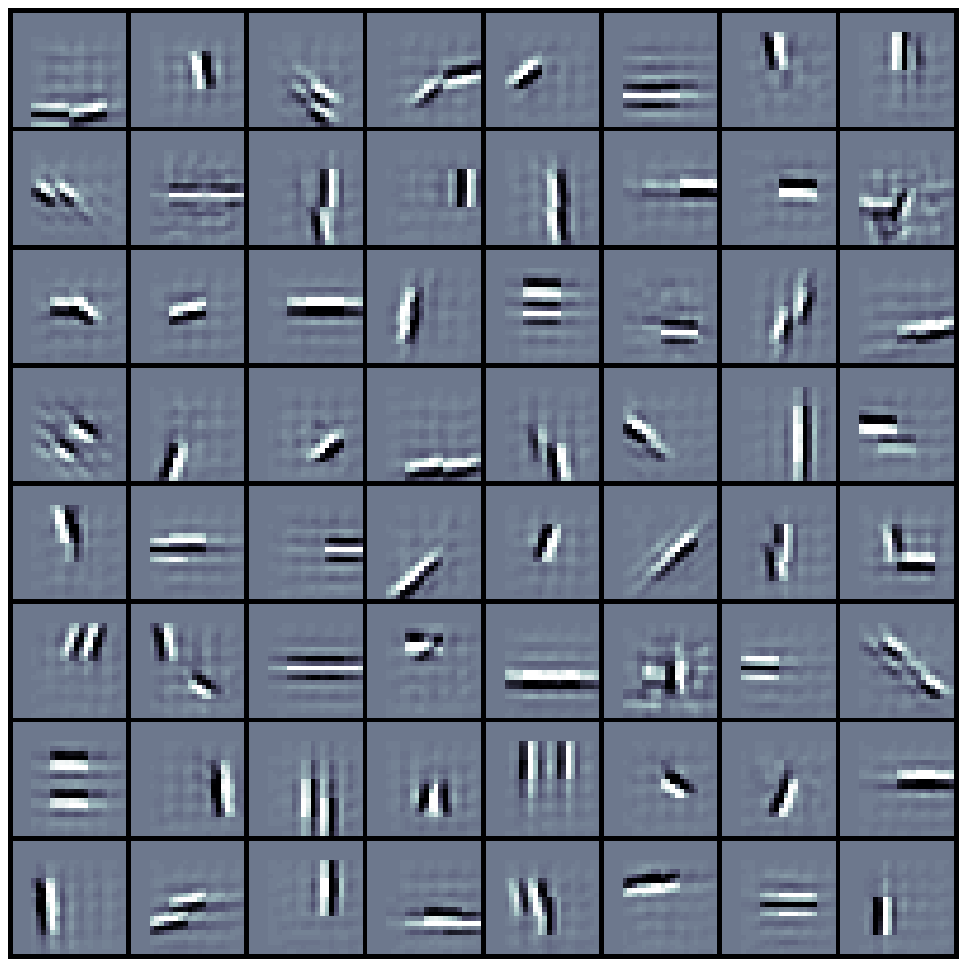}
\end{minipage}
\end{center}
\caption{First and second layer filters from natural images}
\label{natfig}
\end{figure}

We also show results trained on ``unclassified'' natural images from the PASCAL visual object challenge dataset available at \url{http://pascallin.ecs.soton.ac.uk/challenges/VOC/}.  We randomly subsample $5000$ grayscaled images by a factor of 1 to 4, and then pick from each image a $64\times 64$ patch, and then contrast normalize.  We train $8$ $8\times 8$ filters.   We then train a new $4\times 4$ 64 element dictionary on the 
subsampled absolute value rectified responses of the first level.  In figure \ref{natfig} we display the first level filters, and the second level filters up to shifts of size 8 and sign changes of the first level filters.

\begin{figure}
\begin{center}
\begin{minipage}{0.4\textwidth}
\begin{center}
\includegraphics[width=.8\textwidth]{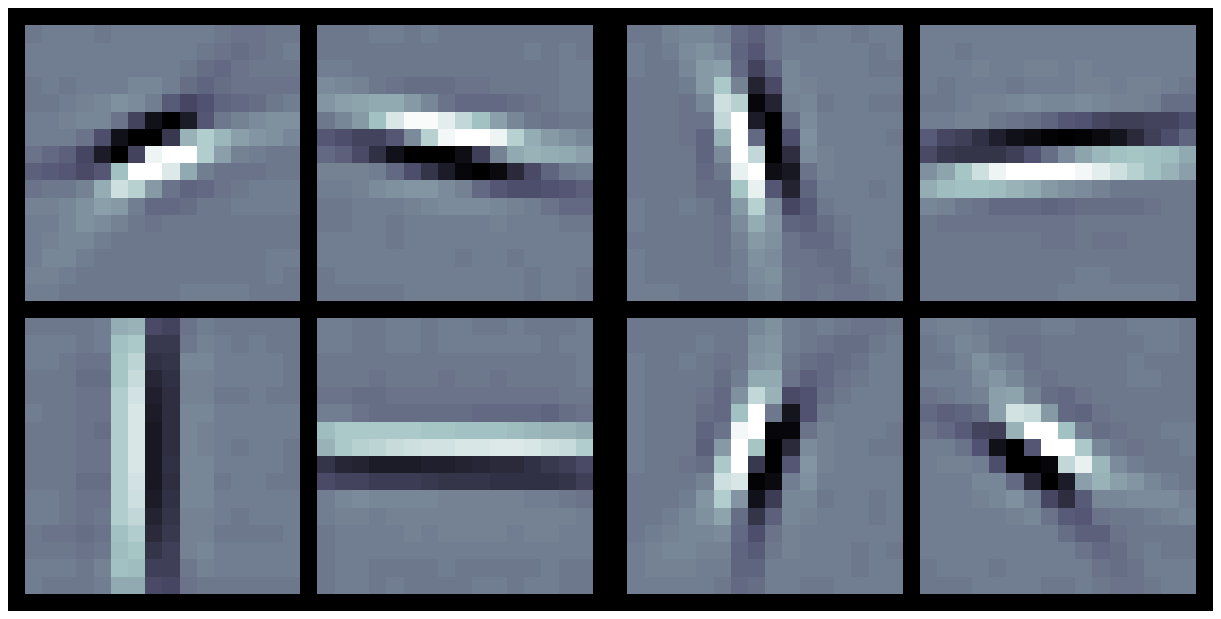}
\includegraphics[width=\textwidth]{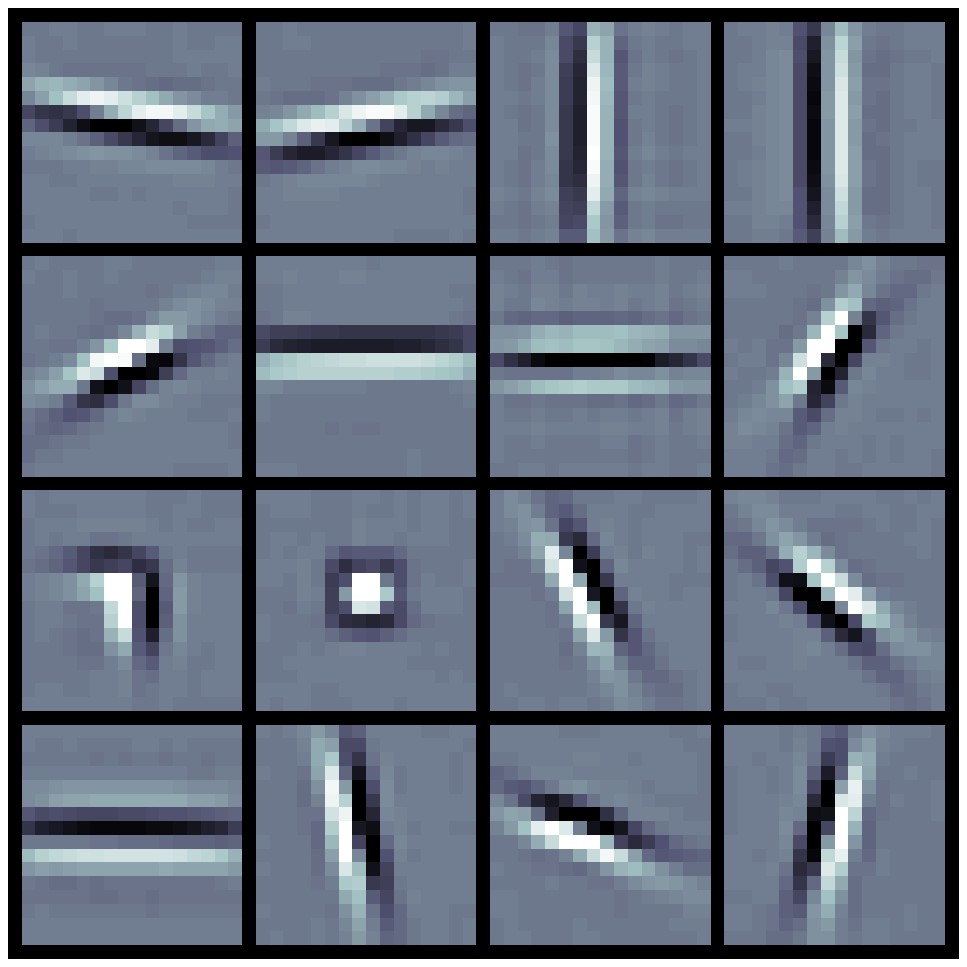}
\end{center}
\end{minipage}
\begin{minipage}{0.8\textwidth}
\includegraphics[width=\textwidth]{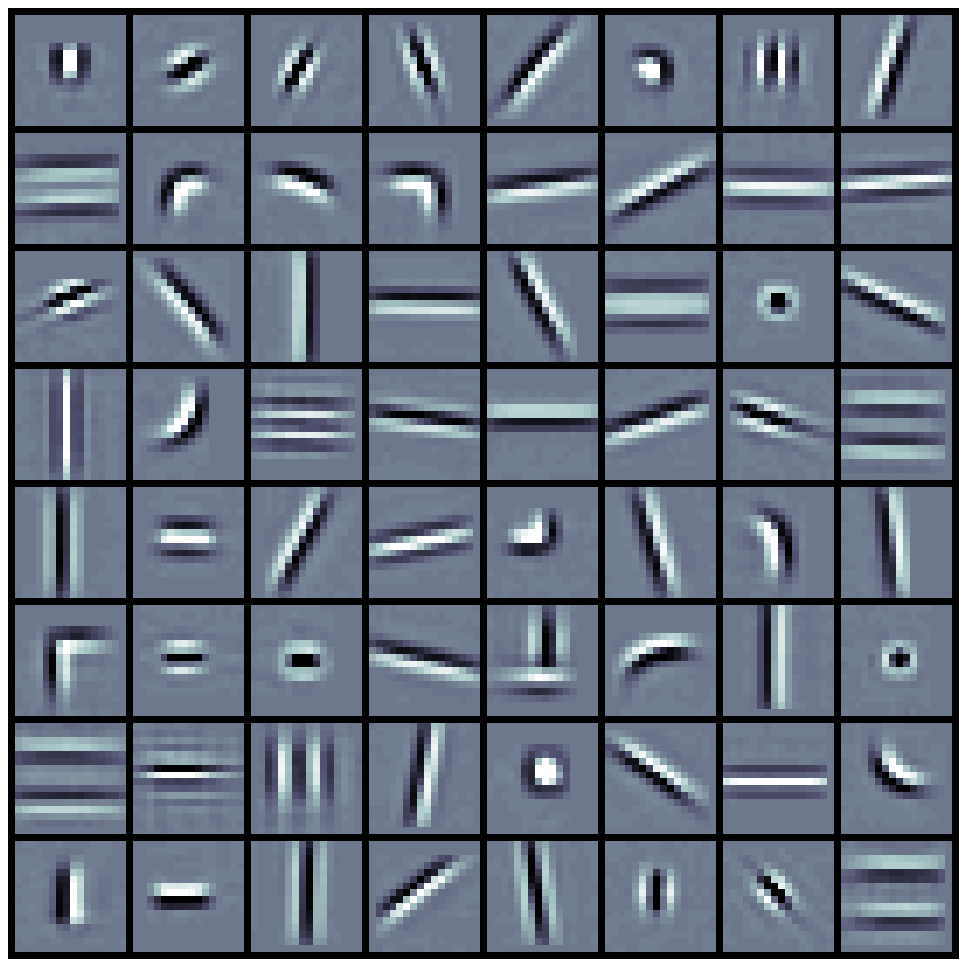}
\end{minipage}
\end{center}
\caption{Dictionaries with varying numbers of elements trained on natural images.}
\label{natfig2}
\end{figure}

In order to show the dependence of the filters on the number of filters used, in figure \ref{natfig2} we display 
an $8$, $16$, and $64$ element $16\times 16$ dictionary trained on the same set as above.

\bibliographystyle{plain}
\bibliography{article}

\begin{thebibliography}{1}

\bibitem{elad-ksvd}
M.~Aharon, M.~Elad, and A.~Bruckstein.
\newblock {K-SVD}: An algorithm for designing overcomplete dictionaries for
  sparse representation.
\newblock {\em IEEE Transactions on Signal Processing}, 54(11):4311--4322,
  2006.

\bibitem{jmlrBach08a}
Francis~R. Bach.
\newblock Consistency of the group lasso and multiple kernel learning.
\newblock {\em Journal of Machine Learning Research}, 9:1179--1225, 2008.

\bibitem{Behnkeconvnnmf}
Sven Behnke.
\newblock Discovering hierarchical speech features using convolutional
  non-negative matrix factorization.
\newblock In {\em IJCNN}, pages 7--12, 2008.

\bibitem{bruckstein:34}
Alfred~M. Bruckstein, David~L. Donoho, and Michael Elad.
\newblock From sparse solutions of systems of equations to sparse modeling of
  signals and images.
\newblock {\em SIAM Review}, 51(1):34--81, 2009.

\bibitem{koraynips2010}
Y.~Boureau K. Gregor M. Mathieu Y.~LeCun K.~kavukcuoglu, P.~Sermanet.
\newblock Learning convolutional feature hierarchies for visual recognition.
\newblock {\em Advances in NIPS}, 2010.

\bibitem{Mallat93matchingpursuit}
Stephane Mallat and Zhifeng Zhang.
\newblock Matching pursuit with time-frequency dictionaries.
\newblock {\em IEEE Transactions on Signal Processing}, 41:3397--3415, 1993.

\bibitem{zeilercvpr2010}
Graham~Taylor Matthew~Zeiler, Dilip~Krishnan and Rob Fergus.
\newblock Hierarchical convolutional sparse image decomposition.
\newblock In {\em The Twenty-Third IEEE Conference on Computer Vision and
  Pattern Recognition}, San Francisco, CA, June 2010.

\bibitem{olshausen97sparse}
B.~Olshausen and D.~Field.
\newblock Sparse coding with an overcomplete basis set: A strategy employed by
  v1?, 1997.

\bibitem{Yuan06modelselection}
Ming Yuan, Ming Yuan, Yi~Lin, and Yi~Lin.
\newblock Model selection and estimation in regression with grouped variables.
\newblock {\em Journal of the Royal Statistical Society, Series B}, 68:49--67,
  2006.

\end{thebibliography}
\end{document}